\documentclass[10pt,twocolumn,letterpaper]{article}

\usepackage[pagenumbers]{cvpr} 

\usepackage{graphicx}
\usepackage{amsmath}
\usepackage{amssymb}
\usepackage{booktabs}
\usepackage{paralist}
\usepackage{paralist}
\usepackage{algorithm}
\usepackage{algorithmicx,algpseudocode}
\usepackage{float}
\usepackage{adjustbox}
\usepackage{multirow}
\usepackage{fancyhdr}
\usepackage[pagebackref,breaklinks,colorlinks]{hyperref}

\usepackage[capitalize]{cleveref}
\crefname{section}{Sec.}{Secs.}
\Crefname{section}{Section}{Sections}
\Crefname{table}{Table}{Tables}
\crefname{table}{Tab.}{Tabs.}

\def\cvprPaperID{*****} 
\def\confName{CVPR}
\def\confYear{2023}

\begin{document}

\title{Zero-shot  Generation of Coherent Storybook  from Plain Text Story using Diffusion Models}

\author{Hyeonho Jeong\\
Department of Software, Sungkyunkwan University\\
{\tt\small drake6751@g.skku.edu}
\and
Gihyun Kwon\\
Department of Bio and Brain Engineering, KAIST\\
{\tt\small cyclomon@kaist.ac.kr}
\and
Jong Chul Ye\\
Kim Jaechul Graduate School of AI, KAIST\\
{\tt\small jong.ye@kaist.ac.kr}
}
\twocolumn[{%
\renewcommand\twocolumn[1][]{#1}%
\maketitle
\begin{center}
\centering
    \includegraphics[width=\textwidth]{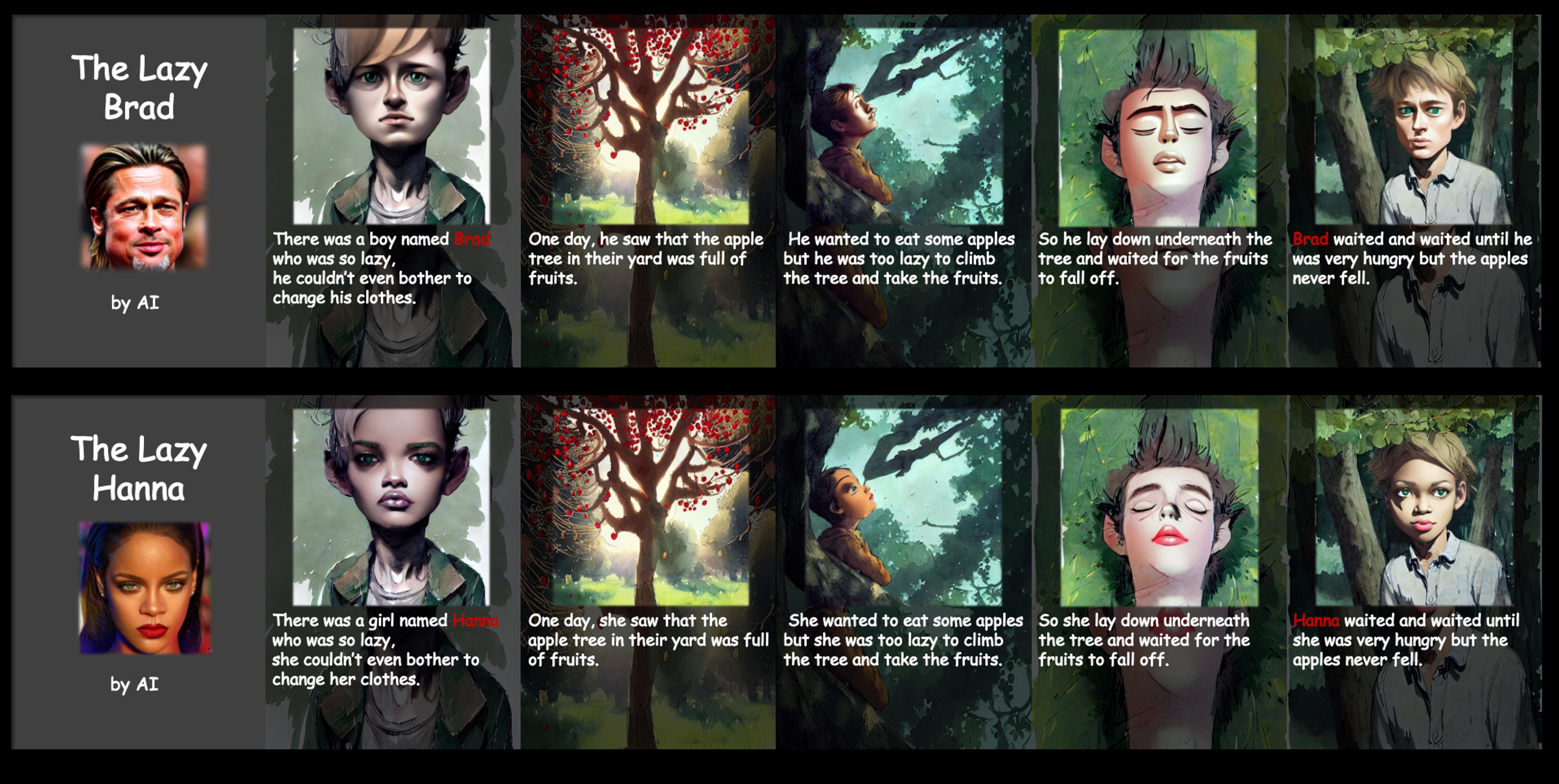}
 \vspace*{-0.2cm}
\captionof{figure}{Zero-shot generation example of a coherent storybook with the main character, Brad (top) and Hanna (bottom), using plain text story from  {\em`The Lazy John'.}
    All the generation processes are performed using large language models (LLM) and latent diffusion models without additional training data.}
\label{teaser-figure} 
\end{center}}]
\begin{abstract}
Recent advancements in large scale text-to-image models have opened new possibilities for guiding the creation of images through human-devised natural language. 
However, while prior literature has primarily focused on the generation of individual images, 
it is essential to consider the capability of these models to ensure coherency within a sequence of images to fulfill the demands of real-world applications such as storytelling.
To address this, here we present a novel neural  pipeline for generating a coherent  storybook from the plain text of a story.
Specifically, we leverage a combination of a pre-trained Large Language Model and a text-guided Latent Diffusion Model to generate 
coherent images.
While previous story synthesis frameworks  typically require a large-scale text-to-image model trained on expensive image-caption pairs to maintain the coherency, 
we employ  simple textual inversion techniques along with detector-based semantic image editing which allows zero-shot generation of the coherent storybook.
Experimental results show that our proposed method outperforms state-of-the-art image editing baselines.

\end{abstract}





\maketitle

\section{Introduction}
The progress in image synthesis using  large-scale text-to-image models has garnered significant attention due to their unprecedented synthetic capacity. 
Text-guided image synthesis methods such as VQGAN-CLIP \cite{crowson2022vqgan} and CLIP-guided Diffusion \cite{crowson2022}, utilize the superior performance of text-to-image alignment in the CLIP \cite{radford2021learning} latent spaces.

Parti \cite{yu2022scaling}, Make-a-scene \cite{gafni2022make}, and DALL-E 1 \cite{ramesh2021zero} have employed large auto-regressive architectures, while Imagen \cite{saharia2022photorealistic}, DALL-E 2 \cite{ramesh2022hierarchical}, and Stable Diffusion \cite{rombach2022high} have utilized diffusion-based architectures. MUSE \cite{chang2023muse} employs a masked transformer architecture and learns to predict masked tokens using the embeddings projected from  Large Language Models (LLM) and image tokens. 
The overwhelming performance of these models has sparked ongoing research efforts to harness their power for 
image editing.

While prior frameworks have mainly focused on the generation of individual, high-quality images, real-world applications place an equal emphasis on coherency within a series of generated images. For example, in the context of an AI-generated storybook, it is crucial for the main character to maintain a consistent appearance throughout the book.
Therefore, we address the challenge of not only locally editing an image, but also applying these techniques to a practical real-world application: the generation of a training-free storybook that maintains the coherency of the main character.

One prior attempt at this is Long Stable Diffusion \cite{long-stable}, which generates a book (both plot and corresponding illustrations) using a combination of LLM and text-image models. However, their use of a LLM is limited to generating illustration ideas, which can lead to irrelevant images if directly used as text conditioning or require significant manual modification of prompts.
Another pioneering work in the field of story synthesis is StoryGAN \cite{li2019storygan}, a conditional GAN \cite{mirza2014conditional} trained to generate visual sequences corresponding to text descriptions of a story.
Subsequently, StoryDALL-E \cite{maharana2022storydall} proposed the adaptation of pre-trained text-to-image transformers to generate visual sequences that continue a given text-based story, introducing a new task of story continuation. More recently, Pan et al. \cite{pan2022synthesizing} proposed history-aware Auto-Regressive Latent Diffusion Models that leverage diffusion models for story synthesis, utilizing a history-aware conditioning network to encode past caption-image pairs. While previous works have made important strides in story synthesis by utilizing datasets of specific domains \cite{zeng2019pororogan} for training, generalizing their methods to other domains or stories often requires expensive image-caption pairs and additional training.

To address these issues, here we introduce  a neural pipeline for the zero-shot generation
of coherent storybooks without additional training data.
Specifically, our method starts with
a simple, yet powerful prompt generation pipeline that takes as input the plain text of existing stories. Furthermore, we ensure that the main character maintains a consistent appearance throughout the book by utilizing our proposed semantic image editing method that injects the desired identity into the facial regions, which play a crucial role in distinguishing a character's identity. Our experimental results demonstrate the effectiveness of our approach in comparison to state-of-the-art semantic image editing baselines.

In summary, the contributions of our work are as follows:
\begin{itemize}
    \item We propose a novel prompt generation pipeline, in which LLMs understand the context and generate prompt inputs for text-to-image models, replacing the need for human-devised natural language prompts.
    
    \item We propose our semantic image editing method and demonstrate its effectiveness against other baselines in terms of smooth editing, coherency maintenance across independent images, and preservation of background regions.
    
    \item Our method allows for fine-grained control over the degree of semantic change by adjusting the number of cycles.
\end{itemize}

\section{Related Works}
\textbf{Large Language Models and Prompt Engineering.}\quad
Large Language Models (LLMs) \cite{devlin2018bert,raffel2020exploring,radford2019language,brown2020language} have gained significant attention in recent years due to their capability to create human-like text and accomplish a wide range of natural language processing (NLP) tasks accurately. 
Recent works demonstrated that LLMs can perform both few-shot \cite{brown2020language} and zero-shot \cite{kojima2022large} reasoning tasks with high accuracy, generalizing well to both common and rare examples. This is achieved by conditioning the models on a few examples or task instructions, a method commonly referred to as "prompting" \cite{liu2021pre}.
While most of the research in prompt engineering has focused on text generation in NLP \cite{shin2020autoprompt, reynolds2021prompt, zhou2022large}, relatively little exploration has been conducted on prompting text generation frameworks in visual tasks, e.g. text-to-image generation.
Various guidelines \cite{liu2022design,oppenlaender2022taxonomy, witteveen2022investigating} for designing prompts have been suggested through empirical analysis of the relationship between prompt components and the resulting visual effects. The studies employed extensive experimentation, utilizing a variety of prompt modifiers on the generation model to gain a deeper understanding of the factors that influence the generated images.

\noindent\textbf{Diffusion Models.}\quad
Diffusion models \cite{sohl2015deep} are trained to learn the underlying data distribution by gradually denoising a variable sampled from Gaussian distribution. In particular, this is equivalent to learning the reverse process of the Markov Chain of fixed length $T$.
In a forward diffusion process $q(x_{t}|x_{t-1})$, noised sampled from Gaussian distribution is added to a ground truth image $x_{0}$ at every time step $t$:
\begin{equation}
\begin{aligned}
q(x_{t}|x_{t-1}) := \mathcal{N}(x_{t};\sqrt{1-\beta_{t}}x_{t-1},\beta_{t}\textbf{I})\\
q(x_{T}|x_{0}) := \prod_{t=1}^{T} q(x_{t}|x_{t-1}),
\end{aligned}
\end{equation}
where $\beta_{t}$ decides the step size which gradually increases. Let $\alpha_{t} := 1 - \beta_{t}$ and 
$\bar{\alpha_{t}} := \prod_{i=1}^{t}\alpha_{i}$. Then we can sample $x_{t}$ in a single step:
\begin{equation}
\begin{aligned}
x_{t} = \sqrt{\bar{\alpha_{t}}}x_{0} +
\sqrt{1-\bar{\alpha_{t}}}\epsilon
\end{aligned}
\end{equation}
where $\epsilon$ is a noise sampled from $\mathcal{N}(0,I)$.
DDPM \cite{ho2020denoising} learns to predict the noise component of $x_{t}$ and generate a marginally denoised $x_{t-1}$ from $x_{t}$ by maximizing the variational lower bound to minimize the negative log-likelihood of $p_{\theta}(x_{0})$. A function $\epsilon_{\theta}$ is parameterized by a UNet-shaped model to reverse the forward step and mean $\mu_{\theta}(x_{t},t)$ is learned in the reverse process:
\begin{equation}
\begin{aligned}
p_{\theta}(x_{t-1}|x_{t}) := \mathcal{N}(x_{t-1}; \mu_{\theta}(x_{t},t), \sigma^{2}_{t}\textbf{I}), \quad \text{where} \\
\mu_{\theta}(x_{t},t) := \cfrac{1}{\sqrt{\alpha_{t}}}
\big(x_{t}-\cfrac{1-\alpha_{t}}{\sqrt{1-\bar{\alpha_{t}}}} \epsilon_{\theta}(x_{t},t)\big)
\end{aligned}
\end{equation}
The training objective of the diffusion model is then:
\begin{equation}
\begin{aligned}
L_{\mathrm{DM}} :=\mathbb{E}_{x_{0}, \epsilon\sim\mathcal{N}(0,1),t}\left[\left\|\epsilon-\epsilon_{\theta}
\left( \sqrt{\hat{\alpha}_{t}}x_{0} + \sqrt{1-\hat{\alpha}_{t}}\epsilon_{t}
,t\right)\right\|^{2}\right]
\end{aligned}
\end{equation}
\\

Denoising Diffusion Implicit Models (DDIM) \cite{song2020denoising} proposed a method for accelerating the sampling process of diffusion models by replacing the Markov forward process used in DDPM with a non-Markovian process.
DDIM formulates a Markov chain that reverses a non-Markovian perturbation process. This chain is fully deterministic when the variance of the reverse noise $\alpha^{2}_{t}$ is set to zero. DDIM method is formulated using the following Markov chain:
\begin{equation}
\begin{aligned}
q(x_{1},...,x_{T}|x_{0}) = \prod_{t=1}^{T} q(x_t | x_{t-1}, x_0)\\
q_\sigma(x_{t-1} | x_t, x_0) := \mathcal{N}(x_{t-1}; \tilde{\mu}_t(x_t, x_0), \sigma_t^2 \textbf{I})
\end{aligned}
\end{equation}
where
\begin{equation}
\begin{aligned}
\tilde{\mu}_t(x_t, x_0) = \sqrt{\tilde{\alpha}_{t-1}} x_0 + \sqrt{1 - \tilde{\alpha}_{t-1} - \sigma_t^2} \cdot\cfrac{x_t - \sqrt{\tilde{\alpha_{t}}}x_0}{\sqrt{1-\tilde{\alpha_t}}}
\end{aligned}
\end{equation}

\noindent\textbf{Text-to-Image Generation and Semantic Image Editing.}\quad
The field of text-guided synthesis has seen a significant amount of progress in recent years, with various methods proposed to address the challenge of generating images from natural language descriptions. 
Early works \cite{mirza2014conditional, reed2016generative, xu2018attngan, zhang2017stackgan,zhang2018stackgan++} in this field were based on Generative Adversarial Networks (GAN) \cite{goodfellow2020generative}. 

A widely adopted approach is the use of auxiliary models such as CLIP \cite{radford2021learning} to guide the optimization of pre-trained generators towards the objective of minimizing text-to-image similarity scores \cite{crowson2022vqgan,crowson2022}.
Additionally, other works have exploited the use of CLIP in conjunction with generative models for various tasks such as image manipulation \cite{patashnik2021styleclip, kim2021diffusionclip}, domain adaptation \cite{gal2022stylegan}, style transfer \cite{kwon2022clipstyler}, and even object segmentation \cite{luddecke2022image,wang2022cris}.
Recently, large-scale text-to-image models demonstrated impressive image generation performance  \cite{yu2022scaling,gafni2022make,ramesh2021zero,ramesh2022hierarchical,saharia2022photorealistic,rombach2022high,chang2023muse}. 
These models have sparked  research efforts using diffusion models   \cite{sohl2015deep} for text-driven image generation and editing. 
In regard to image editing, previous works such as Blended Diffusion \cite{avrahami2022blended1}, Blended Latent Diffusion \cite{avrahami2022blended2}, and Paint by Example \cite{yang2022paint} have addressed the issue of background preservation through user-provided masks.
 Another approach demonstrated the ability to edit synthesized images using text prompts as editing guidance \cite{hertz2022prompt}. DALL-E 2 \cite{ramesh2022hierarchical} also proved its impressive capability in text-guided image editing, commonly referred to as `inpainting'.
 However, when generating images from these models, the identity of a subject is not sustained among images. This is because the diffusion-based generators only use text as a guidance and changing the context in the prompt also alters the appearance.

\section{Method}
In this section, we detail our neural pipeline for generating a training-free storybook that maintains the coherence of the main character.
Specifically, for a given plain text of a story, our algorithm automatically generates matched images and storybook
that maintains the coherence of the main character.
The overall process of our method can be summarized as follows:
\begin{enumerate}
    \item Generate prompts by providing instruction to LLM along with the text of a story.
    \item Utilize a Diffusion-based Text-to-Image model to generate initial images from the set of prepared prompts.
    \item Apply Face Restoration on the initial images.
    \item Find the textual embedding of a specific identity by Textual Inversion learning.
    \item Use the obtained textual embedding to perform Iterative Coherent Identity Injection on enhanced images.
\end{enumerate}

\begin{figure}[ht]
    \centering{\includegraphics[width=\columnwidth]{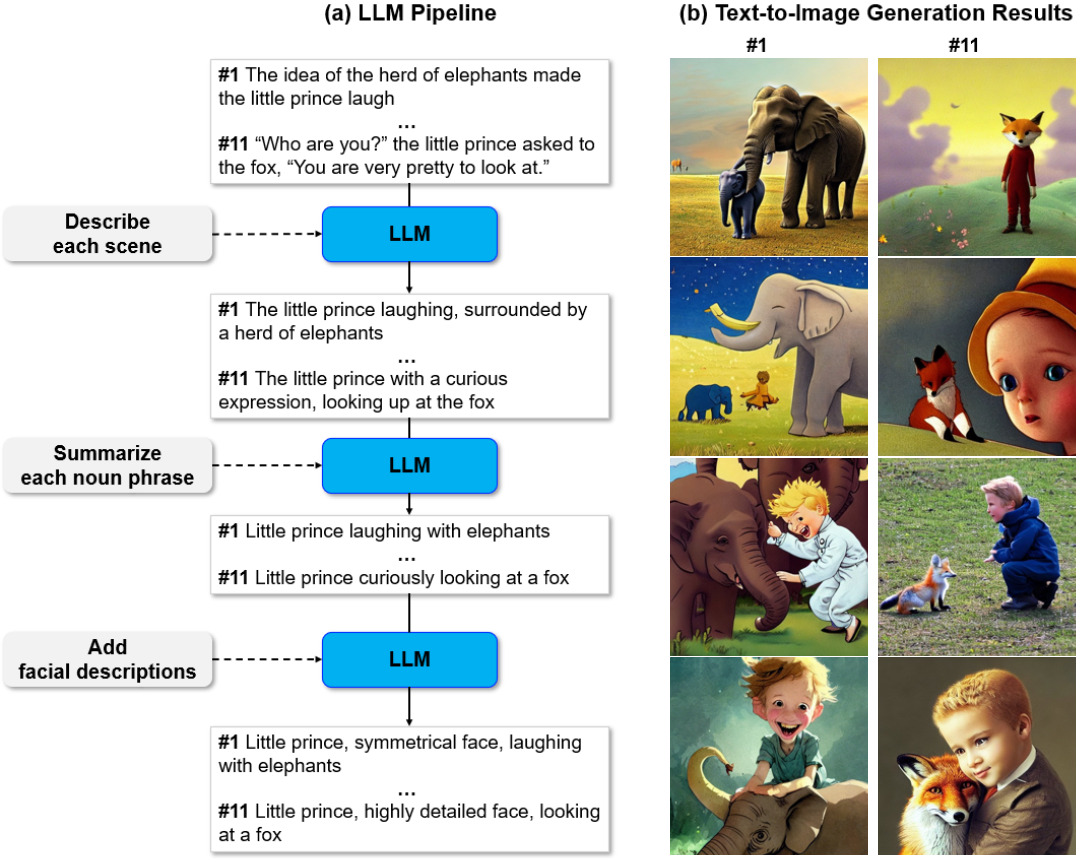}}
    \setlength{\belowcaptionskip}{-10pt}
    \setlength{\abovecaptionskip}{-6pt}
    \caption{(a) Overall prompt generation process. (b) Text-to-Image generation results on the corresponding text sets: it can be observed that the generated images in the lower rows more effectively depict the semantics of the corresponding texts.}
    \label{prompt-pipeline} 
\end{figure}

\begin{figure*}[t]
    \centerline{\includegraphics[width=\textwidth]{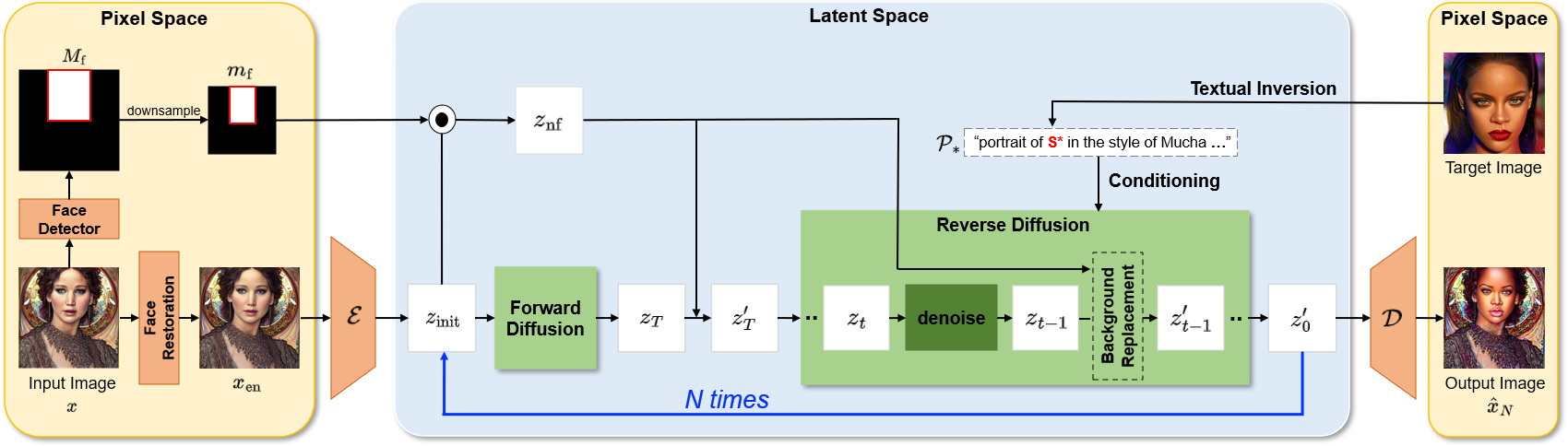}}
    \setlength{\belowcaptionskip}{-10pt}
    \caption{Iterative Coherent Identity Injection procedure.}
    \label{method-illustration} 
\end{figure*}

\subsection{Prompt Generation} 
Prior literature has primarily leveraged text-to-image models for generating a wide range of images with human-devised text prompts.
Instead, this article proposes an approach where we utilize prompts generated by a LLM when given the plain text of a story. 
Our pipeline, as outlined in Figure \ref{prompt-pipeline} (a), is able to generate a series of prompts corresponding to each scene of a story.\\
Let us utilize `The Little Prince' \cite{de2021little} as a case study for our corresponding prompts generation pipeline. 
The process begins by providing the LLM with the instruction to "Describe each scene" along with the sentences of the story.
One example input-output pair for this step would be "The idea of the herd of elephants made the little prince laugh" and "The little prince laughing, surrounded by a herd of elephants".
The LLM expresses its creativity through this step, with the strength of creativity being controlled by the temperature parameter $T$.
However, text-to-image synthesis models often struggle to effectively handle the complexity of the outputs generated from the "Describe each scene" instruction.
As demonstrated in Figure \ref{prompt-pipeline} (b), the amount of information contained in these outputs exceeds the capacity of text-to-image models.
To address this issue, we employ a summarization step in which we provide the LLM with the "Summarize each noun phrase" instruction and the outputs obtained from the previous step.
An example of the output generated in this step, given the previous output "The little prince laughing, surrounded by a herd of elephants," is "The little prince laughing with elephants."

To further enhance the quality of the generated images, we employ additional processing of the prompts by the LLM, through the use of magic words already popular to practitioners.
These magic words, such as `highly detailed' and `insanely intricate' are known to be effective in adding detailed properties to images.
Additionally, when the main subject of the prompt is a person, we also utilize descriptions of facial features, such as `symmetrical face' and `beautiful eyes' to enhance the realism of the generated images. This is accomplished through the use of simple instructions given to the LLM, specifically, "If the main subject of the prompt is a person, add facial descriptions such as `symmetrical face' or `beautiful eyes', where the LLM expresses its creativity again. 

\begin{figure}[H]
    \centerline{\includegraphics[width=\columnwidth]
    {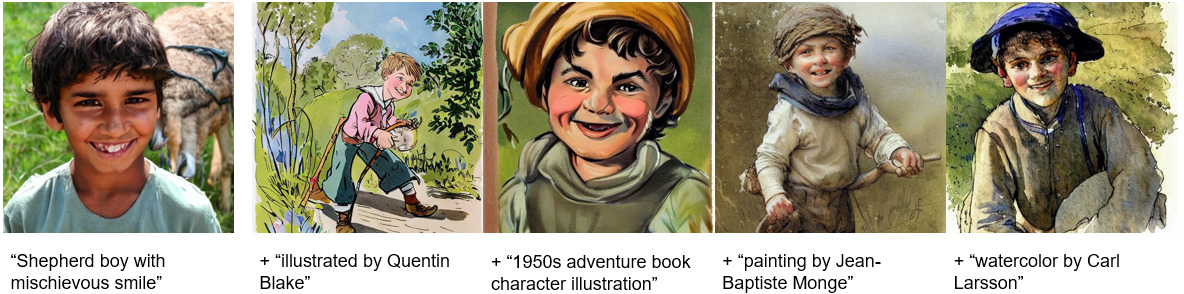}}
    \setlength{\belowcaptionskip}{-6pt}
    \caption{Text-to-image generation with different storybook style modifiers.}
    \label{effects-style-modifiers}
\end{figure}

In order to generate images in a specific style, we employ the use of style modifiers. In this work, we specifically target the style of a storybook. Examples of utilized style modifiers that achieved the storybook aesthetic include:
\begin{inparaenum}
\item Modifiers that include children's book illustrators such as "illustrated by Quentin Blake"
\item Modifiers that indicate a specific type of book such as "1950s adventure book character illustration"
\item Modifiers that include artists with desired artistic style such as "watercolor by Carl Larsson" or "painting by Jean-Baptiste Monge".
\end{inparaenum}
This step of incorporating style modifiers allows for increased scalability as it enables the generation of an infinite number of variations of the storybook by simply altering the style modifiers used, according to the demands of users. Visual effects of mentioned modifiers are illustrated in Figure \ref{effects-style-modifiers}.

\vspace{-2pt}

\vspace{-4pt}
\subsection{Iterative Coherent Identity Injection}
\noindent\textbf{Initial Image Generation using Latent Diffusion Models.}\quad
We utilize a text-conditioned Latent Diffusion model (LDM) \cite{rombach2022high} to generate initial images from a set of prepared prompts. 

Unlike traditional diffusion models that operate in the pixel space, LDM utilizes the latent space, using a pre-trained autoencoder 
(VAE \cite{kingma2013auto} or VQ-VAE \cite{van2017neural,agustsson2017soft}). An encoder $\mathcal{E}$ maps image $x \in \mathbb{R}^{H \times W \times 3}$ to latent representations $z = \mathcal{E}(x), z \in \mathbb{R}^{h \times w \times c}$, while a decoder $\mathcal{D}$ maps the latents back to images $\tilde{x} = \mathcal{D}(z) = \mathcal{D}(\mathcal{E}(x))$. 
Our trained perceptual compression models, the autoencoder consisting of $\mathcal{E}$ and $\mathcal{D}$, enable access to low-dimensional latent space where high-frequency, indiscernible details are ignored. A diffusion model is additionally trained to generate representations within this learned latent space, while conditioned on texts, and initial images. Note that the text condition and the initial image are the two important control knobs that will be used for the iterative coherent identity injection described later.
Based on image-conditioning pairs, the model tries to precisely eliminate the noise added to a latent representation of an image $x$ via:
\begin{equation}
L_{\mathrm{LDM}}:=\mathbb{E}_{z\sim\mathcal{E}(x), y, \epsilon \sim \mathcal{N}(0,1), t}\left[\left\|\epsilon-\epsilon_{\theta}\left(z_{t}, t, \tau_{\theta}(y)\right)\right\|_{2}^{2}\right],
\end{equation}
\noindent where $t$ is the time step, $z_{t}$ is the noisy version of initial latent on time $t$, $\epsilon$ is the sampled noise, $\epsilon_{\theta}$ is a conditional denoising autoencoder, and $\tau_{\theta}$ is a encoder that projects input $y$ to a conditioning vector $\tau_{\theta}(y)$.
$\epsilon_{\theta}$ and $\tau_{\theta}$ are jointly optimized to minimize the loss.

\noindent\textbf{Iterative Coherent Identity Injection.}\quad
Given an initial image $x$ generated by Latent Diffusion Model (LDM) using the prompt $\mathcal{P}$, 
we employ a face restoration model \cite{ wang2021towards,zhou2022towards} on $x$ to obtain an enhanced image $x_{\mathrm{en}}$
 to elevate the quality of the initial images before proceeding to the next step of Iterative Coherent Identity Injection.
This is an important step as it addresses the issue of mutated facial features in the generated images, and also helps to reduce the domain gap between the generated images and the target textual embedding which we obtain from Textual Inversion \cite{gal2022image} of real-photo domain images.
Subsequently, 
we utilize a Face Detector \cite{deng2020retinaface} to extract the bounding box of the facial region in the image.
This bounding box is used to generate a binary mask $M_{\mathrm{f}}$ that marks the region of the face in the image.

To incorporate a target identity, we edit the prompt $\mathcal{P}$ to obtain $\mathcal{P}_{*}$ by replacing the original subject in $\mathcal{P}$ with a placeholder string $S_{*}$ of the target identity, for example replacing "the little prince" in the original prompt with "$S_{*}$".
To find a designated placeholder string $S_{*}$ which  represents the desired target identity given a few images (or a single image) of the target identity, the textual inversion technique proposed by Gal et al.\cite{gal2022image} is used to find the target embedding vector $v_{*}$ in the embedding space of text encoder  $\tau_{\theta}$, where $v_{*}=\tau_{\theta}(S_{*})$. The optimization goal of minimizing the LDM loss is defined as:
\begin{equation}
S_{*}
= \operatorname*{arg\,min}_S \mathbb{E}_{z\sim\mathcal{E}(x), y, \epsilon\sim\mathcal{N}(0,1), t}\left[\left\|\epsilon-\epsilon_{\theta}\left(z_{t}, t, \tau_{\theta}(S)\right)\right\|_{2}^{2}\right]
\end{equation}
where parameters of $\epsilon_{\theta}$ and $\tau_{\theta}$ are frozen.

The objective is then to generate an altered image $\hat{x}_{N}$, where the facial region $x_{\mathrm{en}}\odot M_{\mathrm{f}}$ is consistent to the identity encoded in $S_{*}$, while preserving the background (non-facial) region of the source image. 
The overall $N$ cycles of identity injection process is guided by $\mathcal{P}{*}$ and $M_{\mathrm{f}}$, i.e., LDM performs textual embedding guidance denoising in the latent space of a Variational Autoencoder (VAE) consisting of an encoder $\mathcal{E}$ and a decoder $\mathcal{D}$. 
We denote the facial area as `face' ($\mathrm{f}$) and to the complementary non-facial area as `non-face' ($\mathrm{nf}$). 
The enhanced input $x_{\mathrm{en}}$ first moves to the latent space using the VAE Encoder $z_{\mathrm{init}} \sim \mathcal{E}(x_{\mathrm{en}})$.
As the width and the height of latent representations are downsampled from the input image pixel space, we downsample the facial mask $m_{\mathrm{f}} = \mathrm{downsample}(M_{\mathrm{f}})$ as well. 
With the downsampled facial mask $m_{\mathrm{f}}$, we extract ground truth latent representation of non-face region $z_{\mathrm{nf}}$ by $z_{\mathrm{init}} \odot (1-m_{\mathrm{f}})$.
As the denoising step progresses, the non-facial region of the intermediate representation $z_{t}$ will repeatedly be replaced with the extracted $z_{\mathrm{nf}}$.

More specifically, we do the following process for $N$ cycles:
We noise the initial latent $z_{\mathrm{init}}$ to the noise level $T$. Replace the non-facial region of noised sample $z_{T}$ with the corresponding region of $z_{\mathrm{nf}}$ as 
$z_{T}' \leftarrow z_{T} \odot m_{\mathrm{f}} + z_{\mathrm{nf}}$ . 
The resulting $z_{T}'$ becomes the input of the denoising process.
Then after each denoising step, which is conditioned by the edited prompt $\mathcal{P}_{*}$, we obtain the less noisy sample $z_{t-1}$.
Preserve the facial region of $z_{t-1}$ but replace the complementary region with $z_{nf}$ as
$z_{t-1}' \leftarrow z_{t-1} \odot m_{\mathrm{f}} + z_{\mathrm{nf}}$. 
The resultant $z_{t-1}'$ becomes the input of the next denoising step.
Once the denoising process terminates, instead of being decoded back to pixel space, the resulting denoised sample $z_{0}'$ becomes the input of the next cycle, which eliminates the inefficiency of going back and forth between latent space and pixel space.
After $N$ cycles are carried out, we get the output image by decoding it from the latent space to the pixel space.
The overall procedure is outlined in \textbf{Algorithm 1} and illustrated in Figure \ref{method-illustration}.

\begin{figure*}[h]
    \centerline{\includegraphics[width=\textwidth]{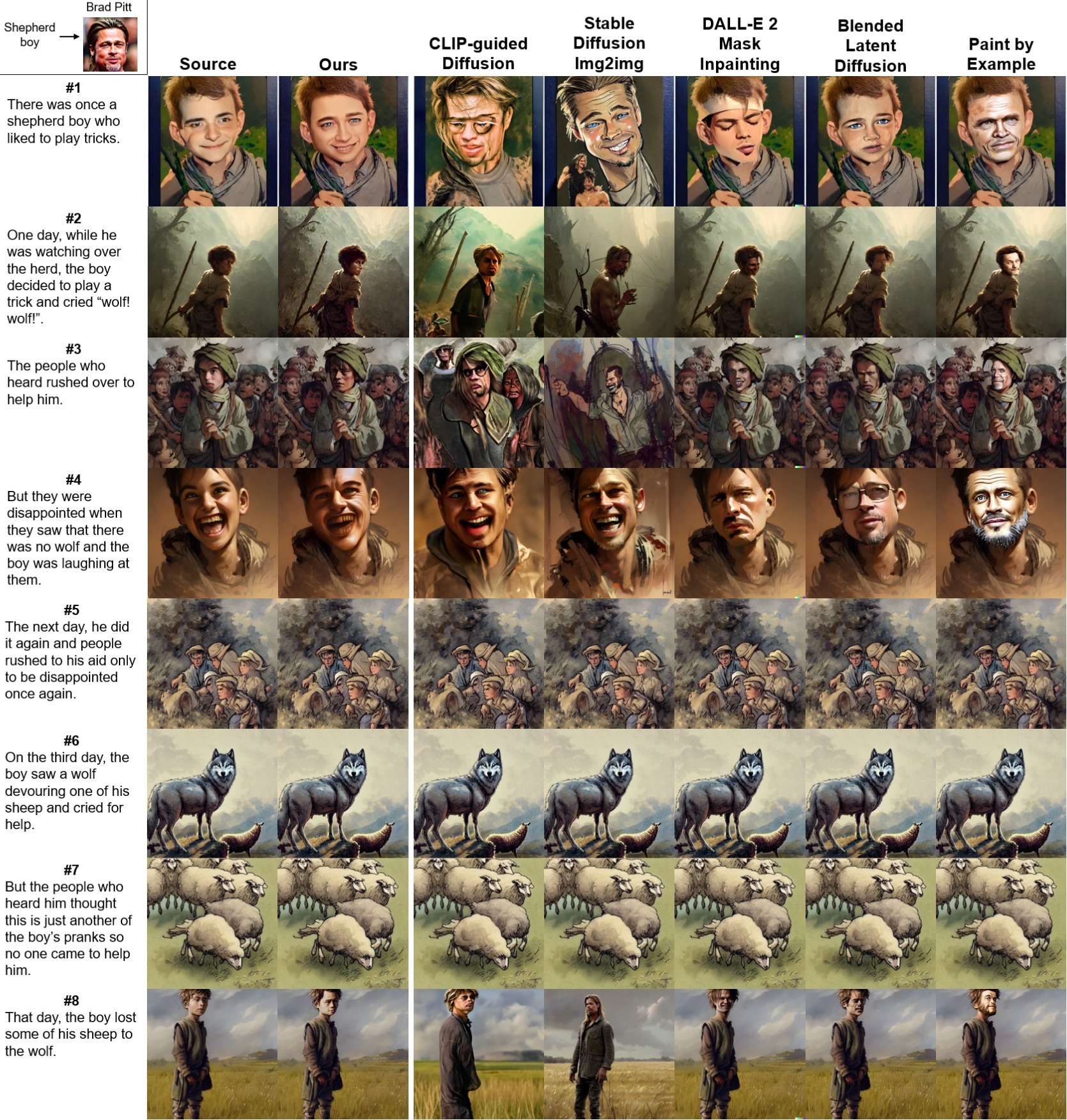}}
    \setlength{\belowcaptionskip}{-10pt}
    \caption{Comparison with semantic image editing baselines. Our method effectively maintains the identity of a character across multiple images while preserving the background of the source images. Note that our method consistently preserves the emotional expressions from the source images, unlike other methods.
    }
    \label{baselines-bradpitt} 
\end{figure*}

\vspace{-2pt}
\begin{algorithm}
\caption{
Iterative Coherent Identity Injection: given a Latent Diffusion Model ($\mathrm{noise}(z,T)$,
$\mathrm{denoise}(z,\mathcal{P},t)$)
along with VAE encoder $\mathcal{E}$ and decoder $\mathcal{D}$
}
\begin{algorithmic}
\State \textbf{Input:} enhanced image $x_{\mathrm{en}}$, modified prompt $\mathcal{P}_*$, downsampled facial mask in latent space $m_{\mathrm{f}}$, the number of cycles $N$, diffusion steps $T$
\State \textbf{Output:} image $\hat{x}_{N}$ that has the appearance of the target identity contained in $S_{*}$, while keeping the background of $x_{\mathrm{en}}$ unchanged
\State
\State $z_{\mathrm{init}} \sim \mathcal{E}(x_{\mathrm{en}})$
\State $z_{\mathrm{nf}} \leftarrow z_{\mathrm{init}} \odot (1-m_{\mathrm{f}})$
\For{$n=1, 2,...,N$}
    \State $z_{T} \sim 
    \mathrm{noise}(z_{\mathrm{init}}, T)$
    \State $z_{T}' \leftarrow z_{T} \odot m_{\mathrm{f}}
    + z_{\mathrm{nf}}$ 
    \For{$t=T,T-1,...,1$}
    \State $z_{t-1} \sim \mathrm{denoise}(z_{t}',\mathcal{P}_{*},t)$
    \State $z_{t-1}' \leftarrow 
    z_{t-1} \odot m_{\mathrm{f}} 
    + z_{\mathrm{nf}}$
    \EndFor
    \State $z_{\mathrm{init}} \leftarrow z_{0}'$
\EndFor
\State $\hat{x}_{N} = \mathcal{D}(z_{0}')$
\State \textbf{Return} $\hat{x}_{N}$
\end{algorithmic}
\end{algorithm}
\vspace{-8pt}

\section{Experiment}

\subsection{Implementation Details}
We first provide details on the implementation choices of our proposed method. For the story-to-prompts component, we employed the GPT-3 model \cite{brown2020language}, specifically the "text-davinci-003" variant, as well as the ChatGPT language model.
For the text-to-image synthesis component, we adopted the Stable Diffusion model(v1.5) \cite{rombach2022high}, a publicly available text-to-image latent diffusion model trained on LAION-5B \cite{schuhmann2022laion} and its default settings.
We utilized CodeFormer\cite{zhou2022towards} for face restoration and RetinaFace face detector \cite{deng2020retinaface} to accurately detect and align faces in the generated images.

\subsection{Experiment Settings}
The experiments were conducted on a GPU of Quadro RTX 6000.
We employed the DDIM scheduler \cite{song2020denoising} for sampling the latent space of the Stable Diffusion model \cite{rombach2022high}. 
In the initial image generation phase, we used 100 reverse diffusion steps, with scale guidance \cite{ho2022classifier} between 7.5 and 10.
A fidelity weight of 0.5 was used during face restoration \cite{zhou2022towards}.
For one cycle of coherent identity injection, we used about 50 forward and reverse diffusion steps.
The number of coherent identity injection cycles varied between 1 and 8. The generated images had a resolution of 512x512 and the latent representations were of size 4x64x64.
Lastly, the story `The Lazy John' presented in Figure \ref{teaser-figure} is from an article presented in \cite{syllabusfy}, `The Boy Who Cried Wolf' in Figure \ref{baselines-bradpitt} is from \cite{ofhsoupkitchen}, and `The Little Prince' in Figure \ref{baselines-naruto} is from \cite{de2021little}.

\subsection{Qualitative Results}
Figure \ref{teaser-figure} illustrates examples of our zero-shot generation of the storybook with two different identities, which clearly
demonstrates that our method can control and maintain the coherency of the stories that are generated without any training data.

In Figure \ref{baselines-bradpitt}, we present a qualitative evaluation of our proposed method in comparison to state-of-the-art semantic image editing baselines.
To demonstrate the effectiveness of our pipeline, we first show initial images generated by our pipeline, which includes prompt generation, initial text-to-image generation, and face restoration.
We then compare our method with a range of baselines, including text-guided methods such as CLIP-guided Diffusion \cite{crowson2022} and Stable Diffusion Img2img \cite{rombach2022high}, DALL-E2 Inpainting \cite{ramesh2022hierarchical}, Blended Latent Diffusion \cite{avrahami2022blended2} and an image-guided method such as Paint by Example \cite{yang2022paint}. 
The `Source' images show that our pipeline successfully synthesizes images corresponding to the plot of the story. The comparison results show that our method outperforms the baselines in terms of coherency preservation and background region preservation. Further comparisons on additional stories are presented in Figure \ref{baselines-lazy-john} and Figure \ref{baselines-naruto}.

\subsection{Quantitative Results}
We created three questionnaires for a total of four stories, with each questionnaire designed to assess different aspects of the generated images. The first questionnaire, labeled as \textbf{Correspondence}, assessed the level of correspondence between the plot of the text and the generated image pairs. The second questionnaire, labeled as \textbf{Coherency}, assessed the coherency of the main characters across images within a story. Lastly, the third questionnaire, labeled as \textbf{Smoothness}, assessed how smooth and seamless the transition was between the foreground and background in the images. To evaluate detailed opinions, we employed a custom-made opinion scoring system. The results of this study on 76 random people are summarized in Table 1 in the form of mean scores for each of the questionnaires. Overall, our method achieved the highest scores in all aspects, indicating the effectiveness of our proposed pipeline.

\begin{table}[h]
\centering
\begin{adjustbox}{width=0.48\textwidth}
\begin{tabular}{@{\extracolsep{5pt}}c|cccccc@{}}

\multirow{2}{*}{\textbf{Methods}} & {CLIP-guided} & {Stable} & \multirow{2}{*}{DALL-E2} & Blended & {Paint by} & \multirow{2}{*}{Ours}\\

& Diffusion & Diffusion & & Latent Diffusion & Example& \\

\hline
Correspondence& 2.96 & 2.68 & 2.75 & 2.42 & 2.32 & \textbf{4.06} \\
Coherence & 2.16 & 2.64 & 2.44 & 2.36 & 2.24 & \textbf{3.84} \\
Smoothness & 2.55 & 2.87 & 2.96 & 2.71 & 2.20 & \textbf{4.23} \\
\hline

\end{tabular}
\end{adjustbox}
\caption{User study results on various semantic image editing models. Our model outperforms baselines in the user study. }

\label{table:user}
\end{table}

\subsection{Ablation Study}
\begin{figure}[h]
    \centerline{\includegraphics[width=0.9\columnwidth]{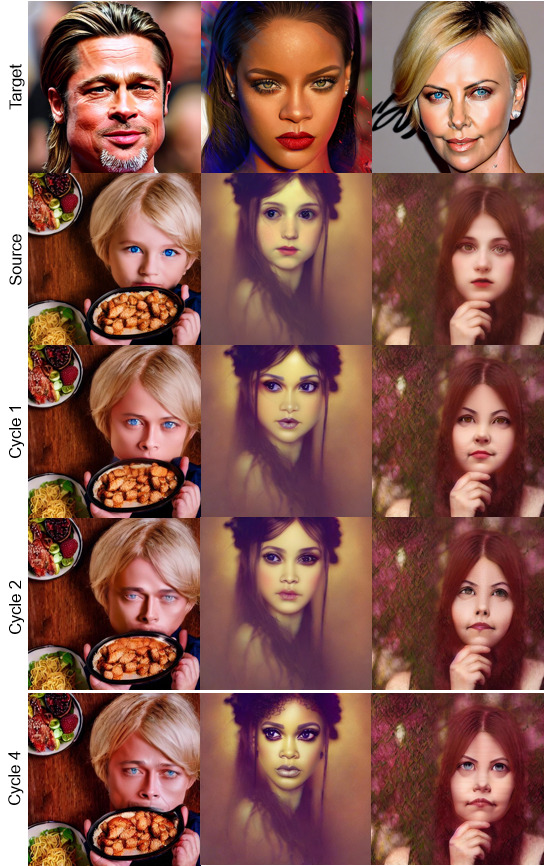}}
    \setlength{\belowcaptionskip}{-7pt}
    \caption{Generation results of our method with a different number of cycles: As the number increases, the injected target identity becomes more apparent in the output image.
    }
    \label{cycles} 
\end{figure}

\begin{figure}[h]
    \centerline{\includegraphics[width=0.9\columnwidth]{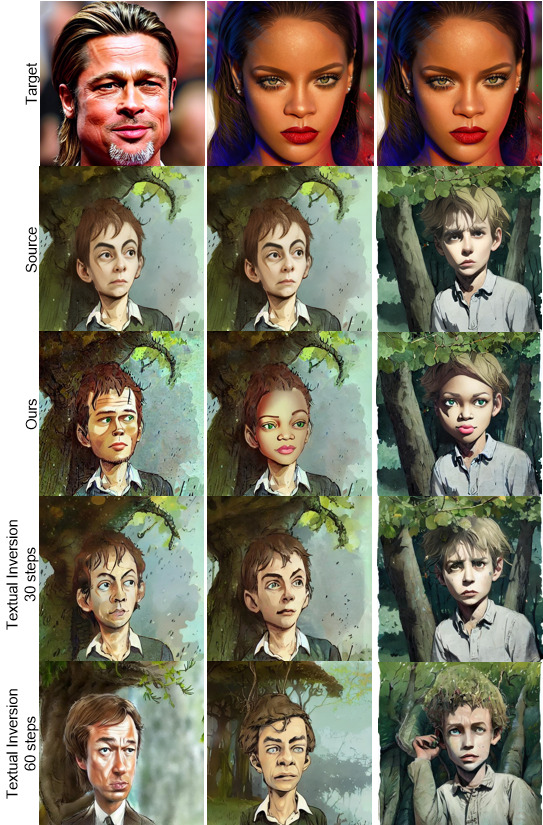}}
    \setlength{\belowcaptionskip}{-8pt}
    \caption{
    Comparison with Textual Inversion method \cite{gal2022image}: Textual embedding guided image synthesis fails to preserve the background of the source images.
    }
    \label{textual-inversion} 
\end{figure}

To validate the optimality of our approach, we performed ablation studies. First, we compare the generative performance on varying numbers of cycles.
The results are presented in Figure \ref{cycles}, which illustrates that as the number of injection cycles increases, the source identity gradually fades away while the target identity becomes more discernible. It demonstrates the ability of the proposed method to provide delicate control over the amount of editing applied to an image. The edited results in the first and the third columns demonstrate that our method is also robust to occlusions.
Additionally, a comparison with a method using only Textual Inversion is presented in Figure \ref{textual-inversion}. It shows that our method is more effective in preserving the background, while the transition between foreground and background appears more seamless.
Further results are presented in Figure \ref{appendix-cycle}.

\section{Conclusion}
In this paper, we presented a new approach for zero-shot storybook synthesis. Our pipeline utilized Large Language Models to generate text-based conditioning, which can replace human-crafted natural language prompts and guide image synthesis models. We then developed an iterative identity injection step, using a textual embedding and a face detector to guide the generation process of Latent Diffusion Models in terms of semantic changes and background preservation. Our experimental results demonstrated that our proposed framework outperforms both text- and image-guided semantic image editing baselines in terms of coherency preservation and background preservation.


{\small
\bibliographystyle{ieee_fullname}
\bibliography{egbib}
}

\appendix

\section{Implementation Details}
In our proposed method, we utilized a large language model, specifically the GPT-3 model \cite{brown2020language} with the "text-davinci-003" version, in conjunction with the ChatGPT language model, to generate prompts for the text-to-image component. Our experiments revealed that both models produce comparable results.
During the story-to-prompts process, the GPT-3 model was configured with a `Temperature' of 0.5 and a `Top P' parameter of 1.
An example instruction set used in our implementation was "Prompts to generate painting that best describes each scene of a story." followed by "Summarize each noun phrase." and lastly, "If the main subject of the prompt is categorized as a human, add facial descriptions on the subject such as symmetrical face',emerald eyes'."

For the initial text-to-image synthesis component, we utilized the official codebase of the Stable Diffusion model \footnote{\url{https://github.com/CompVis/stable-diffusion}} \cite{rombach2022high}
with its default hyperparameter settings and the according model checkpoints from Hugging Face \footnote{\url{https://huggingface.co/runwayml/stable-diffusion-v1-5}}.
We used Textual Inversion method \cite{gal2022image} to find the target textual embedding, referencing both the official codebase \footnote{\url{https://github.com/rinongal/textual_inversion}} and an unofficial codebase \footnote{\url{https://github.com/nicolai256/Stable-textual-inversion_win}} by nicolai256. Additionally, we applied the face restoration step on the initially generated images by using the Codeformer Face Restoration model \footnote{\url{https://github.com/sczhou/CodeFormer},\url{https://huggingface.co/spaces/sczhou/CodeFormer}} \cite{zhou2022towards}, where `Fidelity' parameter was set to 0.5.
To precisely detect and align facial parts in the generated images, we employed the RetinaFace Face Detector model \footnote{\url{https://github.com/biubug6/Pytorch_Retinaface}} \cite{deng2020retinaface}.
Details of the Iterative Coherent Identity Injection method are explained in Section 4.2 of the main paper.

\vspace{-0.2cm}
\section{Experimental Details of Baselines}
\begin{figure}[h]
    \centering{\includegraphics[width=0.8\columnwidth]{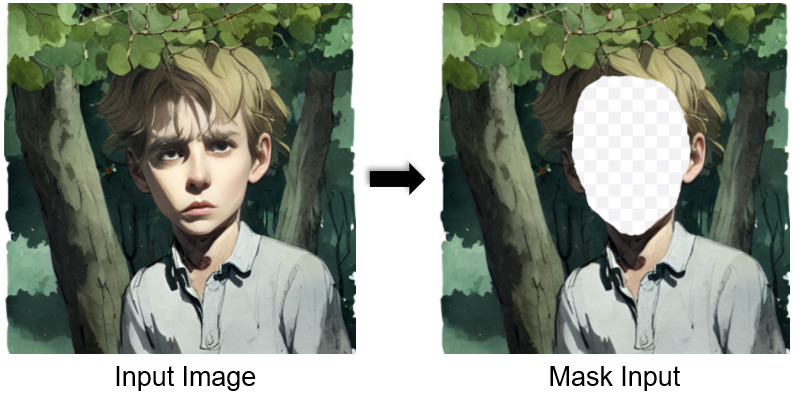}}
    \setlength{\belowcaptionskip}{-15pt}
    \caption{Example of manual masking applied to an input image.}
    \label{masking} 
\end{figure}

In our experiments, we compared our method to several state-of-the-art baselines, including CLIP-guided Diffusion \cite{crowson2022}, Stable Diffusion \cite{rombach2022high}, DALL-E 2 \cite{ramesh2022hierarchical}, Blended Latent Diffusion \cite{avrahami2022blended2}, and Paint by Example \cite{yang2022paint}.
We utilized the official codebase of the works
\footnote{\url{https://github.com/afiaka87/clip-guided-diffusion}}
\footnote{\url{https://github.com/omriav/blended-latent-diffusion}}
\footnote{\url{https://github.com/Fantasy-Studio/Paint-by-Example}}
and according demo web sites \footnote{\url{https://huggingface.co/spaces/EleutherAI/clip-guided-diffusion}} \footnote{\url{https://huggingface.co/spaces/Fantasy-Studio/Paint-by-Example}} if provided. For DALL-E 2, we utilized the paid inpainting demo \footnote{\url{https://openai.com/dall-e-2/}} by OpenAI as we could not access the trained model freely.
In particular, for the experiment with the CLIP-guided Diffusion model, we configured the model with the following hyperparameter settings: {`skip timesteps':30, `clip guidance scale':1600, `tv scale':150, `range scale':50, `initial scale': 1000}. This deviated from the default settings reported in the CLIP Guided Diffusion code repository, as the default settings resulted in poor image quality. 
Additionally, for the methods that are guided by user-provided masks (DALL-E 2, Blended Latent Diffusion, Paint by Example), we manually generated the masks as illustrated in Figure \ref{masking}.

\section{User Study Details}
In order to evaluate the performance of our proposed method, a user study was conducted. A total of 76 participants were recruited, with a diverse age range of the 20s and 30s. The study was conducted using a Google Form, in which participants were asked to provide scores for images generated by our method and several state-of-the-art methods. The evaluation was conducted in a blind manner, in which participants were not informed of which method was used to generate the images.
Three questions were asked for each of the four stories, with each story featuring variations of 6 different image editing methods (1 ours + 5 baselines). 
In total, 72 questions were asked.
The first question assessed the \textbf{Correspondence} of the generated story set, with the following instruction used: "Your task is to evaluate the correspondence between the plot and images, which refers to how well the images depict the corresponding plot."
The second question assessed the \textbf{Coherency} of the generated story set, with the following instruction used: "Your task is to evaluate the coherency of appearance, which refers to how consistent the appearance of the main character is across the images."
The third question assessed the \textbf{Smoothness} of the generated story set, with the following instruction used:
"Your task is to evaluate the smoothness of the images, which refers to how well the facial regions blend seamlessly with the background regions."
A 5-point Likert scale was used for each aspect, with a minimum score of 1 and a maximum score of 5. The options provided were: 1- Very Bad, 2- Bad, 3- Medium, 4-Good, 5-Very Good.

\vspace{-0.2cm}
\section{Limitations}
In our work, we employed a publicly available text-guided Latent Diffusion Model, Stable Diffusion, as the backbone of our proposed method. However, due to its inherent limitations, not all text conditioning methods generated optimal images, particularly in cases where images of people were involved. For example, images with mutated facial or body features were generated in some instances. To address this limitation, we incorporated the step of `Add facial descriptions' in our prompt generation pipeline, which partially mitigated this issue. However, it is important to note that the imperfect nature of Stable Diffusion may still lead to suboptimal results in some cases.

\vspace{-0.2cm}
\section{Social Impact}
The use of Text-Image models and generative image editing techniques poses several ethical challenges.
These models have the potential to be used for malicious purposes, such as creating misleading or fake images, and can have negative societal impacts \cite{fried2020editing}.
Our work is built on such models and is also susceptible to these issues.
Additionally, the models were trained on a dataset of images collected from the web \cite{schuhmann2022laion}, which may contain inappropriate content and biases.
As a result, the models may inherit these biases \cite{mishkin2022dall} and generate inappropriate images.
To mitigate these risks, we will release our code under a license that promotes ethical and legal use, similar to the license used for the Stable Diffusion model.

\section{Additional Results}
\subsection{Face Restoration}
\vspace{-0.2cm}
\begin{figure}[H]
    \centerline{\includegraphics[width=0.9\columnwidth]
    {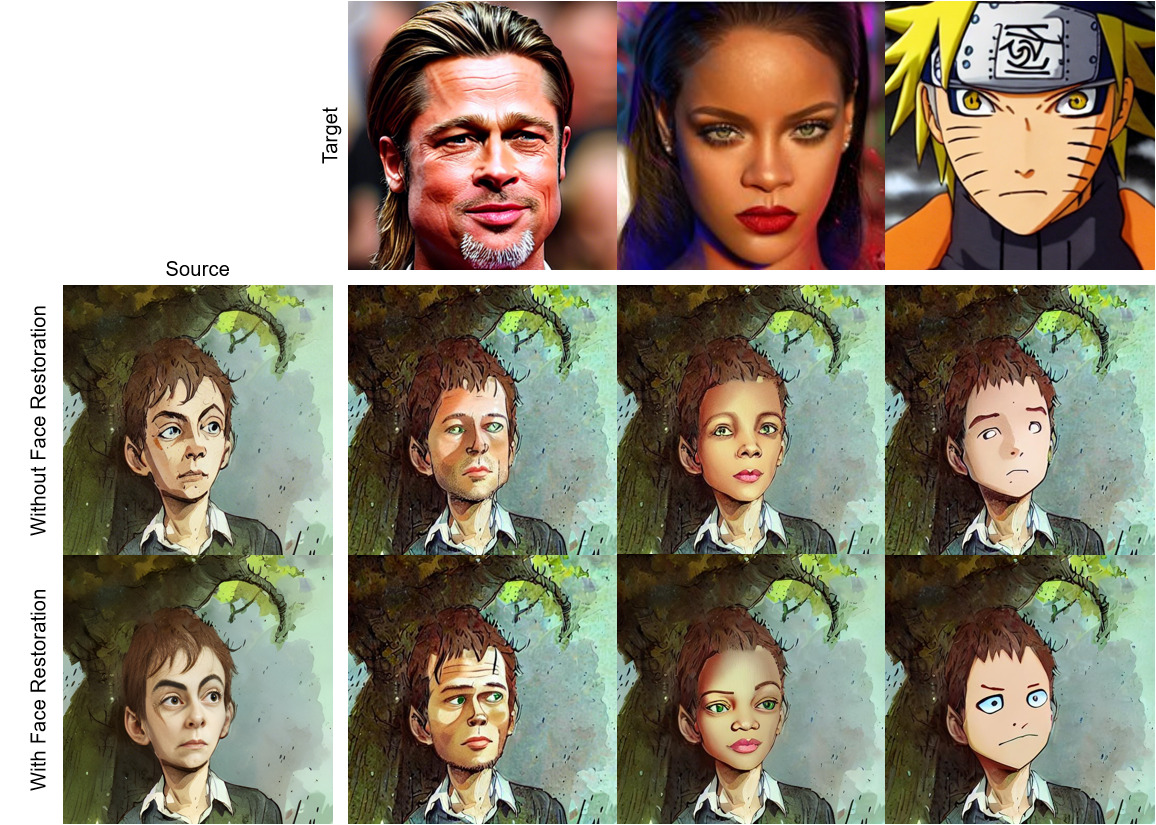}}
    \setlength{\belowcaptionskip}{-3pt}
    \caption{Results of our method with and without face restoration.}
\end{figure}

\subsection{Different Number of Cycles}
\vspace{-0.2cm}
\begin{figure}[H]
    \centerline{\includegraphics[width=0.9\columnwidth]
    {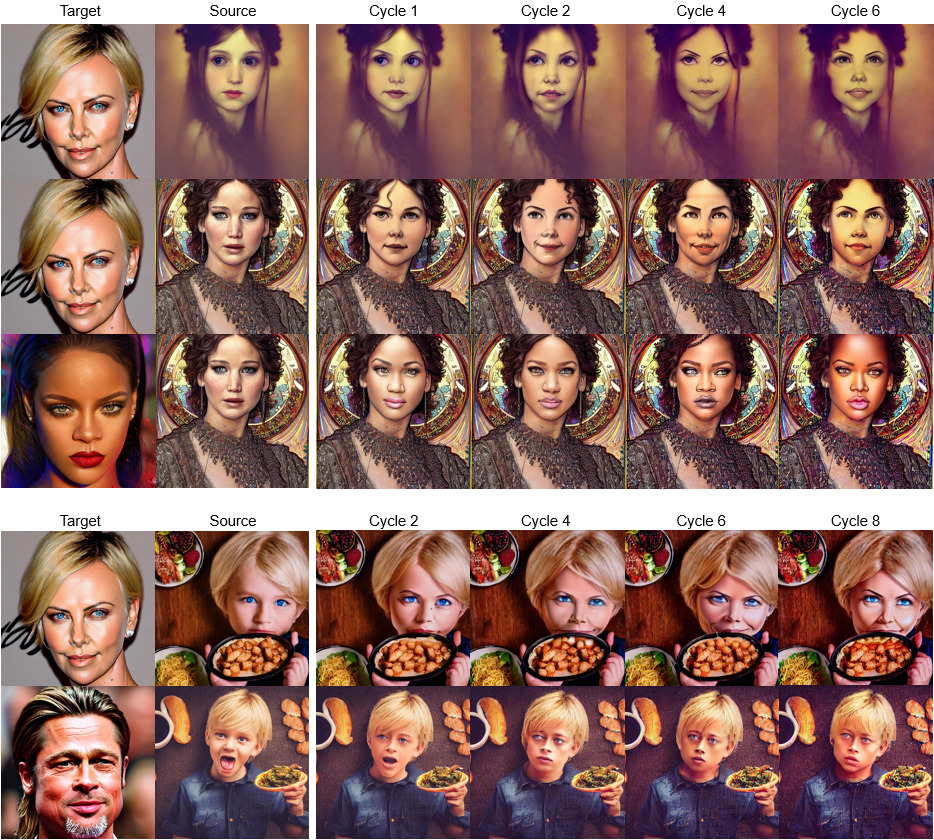}}
    \setlength{\belowcaptionskip}{-10pt}
    \caption{Generation results of our proposed method with different number of cycles}
    \label{appendix-cycle}
\end{figure}

\subsection{Additional Story Synthesis Results}
\vspace{-0.2cm}
\begin{figure}[H]
\centerline{\includegraphics
[width=\columnwidth]
    {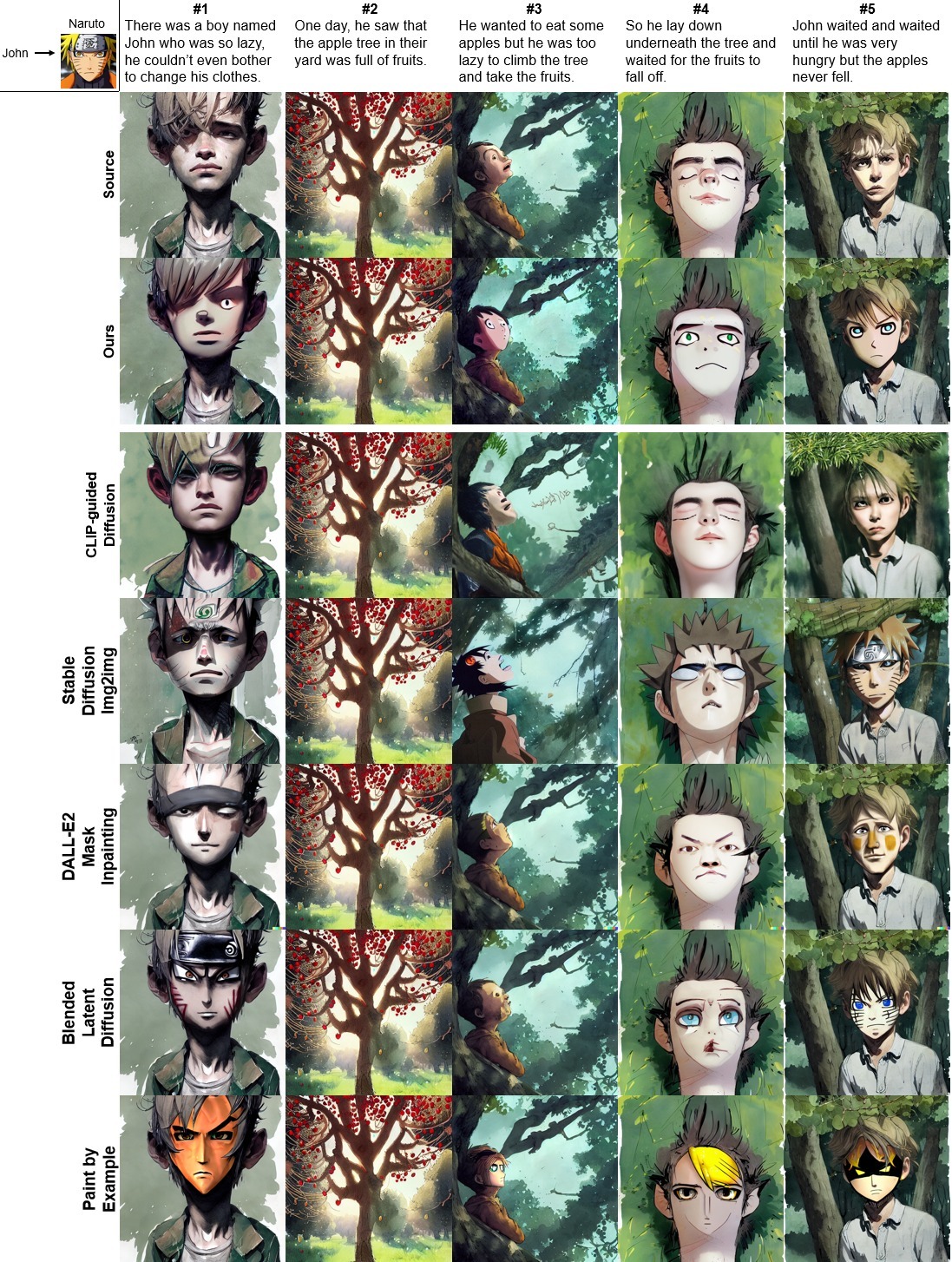}}
    \caption{Our method applied to the story `The Lazy John'}
    \label{baselines-lazy-john}
\end{figure}

\begin{figure*}[]
    \centerline{\includegraphics[width=\textwidth]{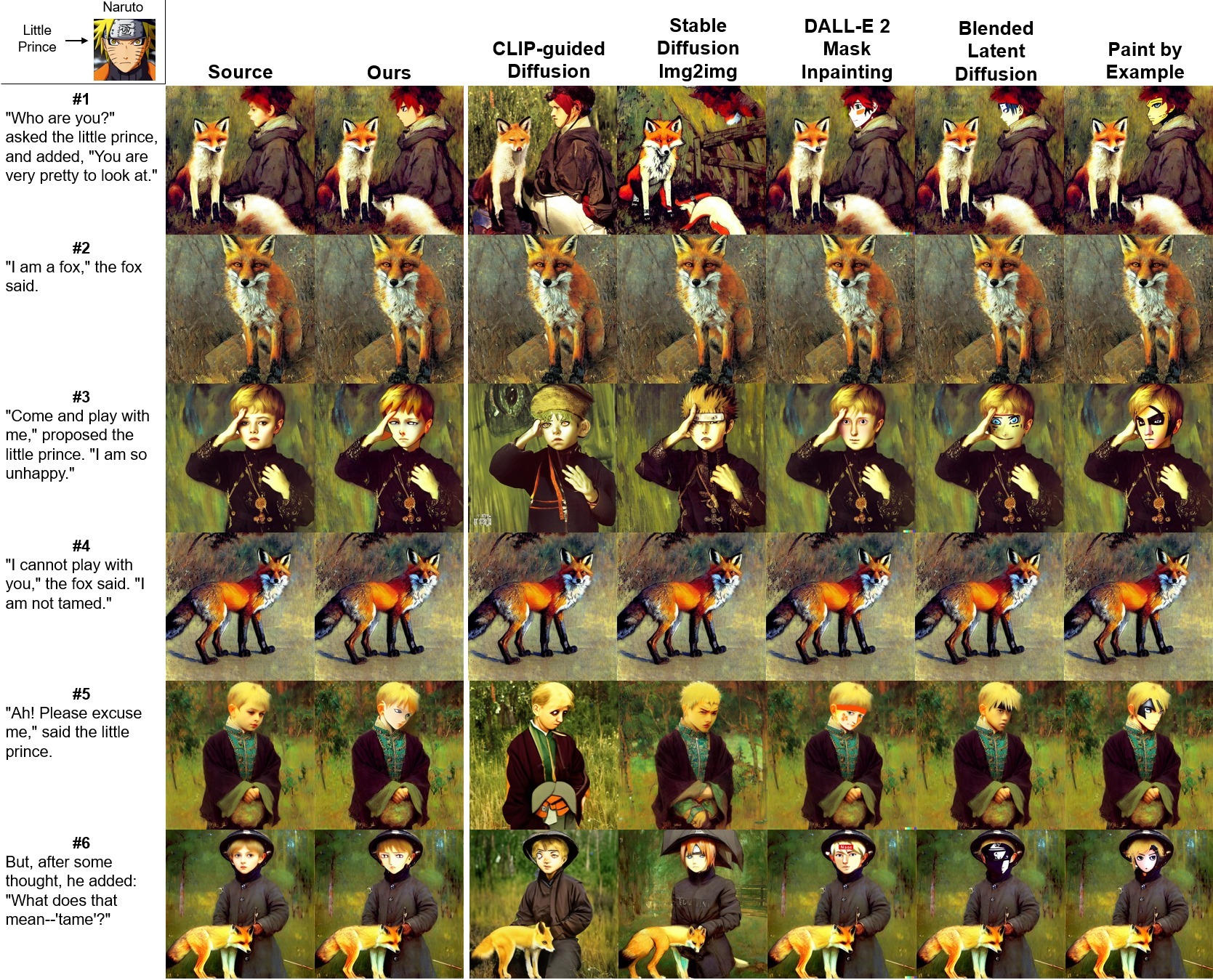}}
    \caption{Our method applied to the story `The Little Prince'}
    \label{baselines-naruto} 
\end{figure*}


\end{document}